\documentclass[letterpaper, 10 pt, conference]{ieeeconf}  

\IEEEoverridecommandlockouts                              

\usepackage{amsmath}
\usepackage{amssymb}
\usepackage{graphicx}
\usepackage{float}
\usepackage{array}
\usepackage{tikz}
\usepackage{listings}
\usepackage{mathrsfs}
\usepackage{pgfplots}
\usepackage{python}
\usepackage[pdfencoding=auto,psdextra,pagebackref,breaklinks,colorlinks]{hyperref}
\usepackage{graphicx}
\usepackage{subcaption}
\usepackage{cite}
\usepackage{dsfont}
\usepackage{mathtools,etoolbox}
\usepackage{algorithm}
\usepackage{algpseudocode}

\usepackage[none]{hyphenat}
\usepackage[utf8]{inputenc}
\usepackage[english]{babel}

\usepackage{amsthm}
\usepackage{xfrac}
\usepackage{nicefrac}
\usepackage{xcolor}
\usepackage{booktabs}
\usepackage[switch, pagewise]{lineno}
\usepackage{nicefrac}
\usepackage{flushend}
\usepackage{gensymb}
\usepackage{lipsum}
\usepackage{balance}
\usepackage{booktabs}
\usepackage{multirow}
\usepackage{xcolor}
\usepackage{dirtytalk}

\usepackage{pgfplots}
\pgfplotsset{compat=1.18}

\DeclareUnicodeCharacter{2212}{-}
\usepackage{caption} 

\title{\LARGE \bf
Post Fusion Bird's Eye View Feature Stabilization for Robust Multimodal 3D Detection
}

\author{Trung Tien Dong$^{1}$, Dev Thakkar$^{1}$, Arman Sargolzaei$^{1}$ and Xiaomin Lin$^{1}$* 
\thanks{*This work was supported by the DEVCOM Army Research Laboratory under CRADA 22-0034-J003 for Efficient Multimodal Learning and by the NVIDIA Academic Grant Program.}
\thanks{*Corresponding Author}
\thanks{$^{1}$College of Engineering, University of South Florida, Tampa, FL 33620, USA
        {\tt\small dongt@usf.edu; devthakkar@usf.edu; armans@usf.edu; xlin2@usf.edu}}%
}

\begin{document}

\maketitle
\thispagestyle{empty}
\pagestyle{empty}

\begin{abstract}
Camera-LiDAR fusion is widely used in autonomous driving to enable accurate 3D object detection. However, bird's-eye view (BEV) fusion detectors can degrade significantly under domain shift and sensor failures, limiting reliability in real-world deployment. Existing robustness approaches often require modifying the fusion architecture or retraining specialized models, making them difficult to integrate into already deployed systems.

We propose a \textbf{Post Fusion Stabilizer (PFS)}, a lightweight module that operates on intermediate BEV representations of existing detectors and produces a refined feature map for the original detection head. The design stabilizes feature statistics under domain shift, suppresses spatial regions affected by sensor degradation, and adaptively restores weakened cues through residual correction. Designed as a near-identity transformation, PFS preserves performance while improving robustness under diverse camera and LiDAR corruptions. Evaluations on the nuScenes benchmark demonstrate that PFS achieves state-of-the-art results in several failure modes, notably improving camera dropout robustness by +1.2\% and low-light performance by +4.4\% mAP while maintaining a lightweight footprint of only 3.3~M parameters.

\end{abstract}
\section{INTRODUCTION}
Reliable 3D perception is a crucial prerequisite for the safe deployment of autonomous vehicles (AVs) in unstructured traffic environments. Modern robotic perception stacks fuse complementary modalities, typically surround-view cameras and spinning LiDAR, to provide a consistent world-model required for safe motion planning \cite{huang2024multimodalsensorfusionauto}. Among the many paradigms, bird's-eye view (BEV) fusion has emerged as the dominant framework for camera-LiDAR 3D object detection in autonomous driving
as it aggregates multi-sensor features within a unified ego-centric spatial grid aligned with downstream planning tasks.
\cite{ping2025comprehensive}. This representation typically relies on \say{lifting} multi-view image features into 3D using learned depth \cite{philion2020liftsplatshootencoding}, a strategy refined by recent architectures like BEVDet \cite{huang2022bevdethighperformancemulticamera3d}, BEVDepth \cite{li2022bevdepthacquisitionreliabledepth}, and BEVFormer \cite{li2022bevformerlearningbirdseyeviewrepresentation} to establish state-of-the-art results on large-scale benchmarks such as nuScenes \cite{nuscenes2019} and Waymo \cite{Sun_2020_CVPR}.

Despite rapid progress in BEV fusion, multimodal detectors remain brittle under distribution shift. Real deployments face appearance changes, partial observations, and inter modal misalignment. MetaBEV \cite{ge2023metabevsolvingsensorfailures} improves robustness with modality arbitrary decoding, and MoME \cite{park2025resilientsensorfusionadverse} uses mixture of expert decoders with adaptive routing to rely on the most reliable sensors. However, many approaches require architectural changes or backbone replacement, which is costly to validate and calibrate in tightly coupled autonomous driving stacks.


\begin{figure}[t]
    \centering
    \includegraphics[width=0.9\columnwidth, trim=0 5cm 22.3cm 0, clip]{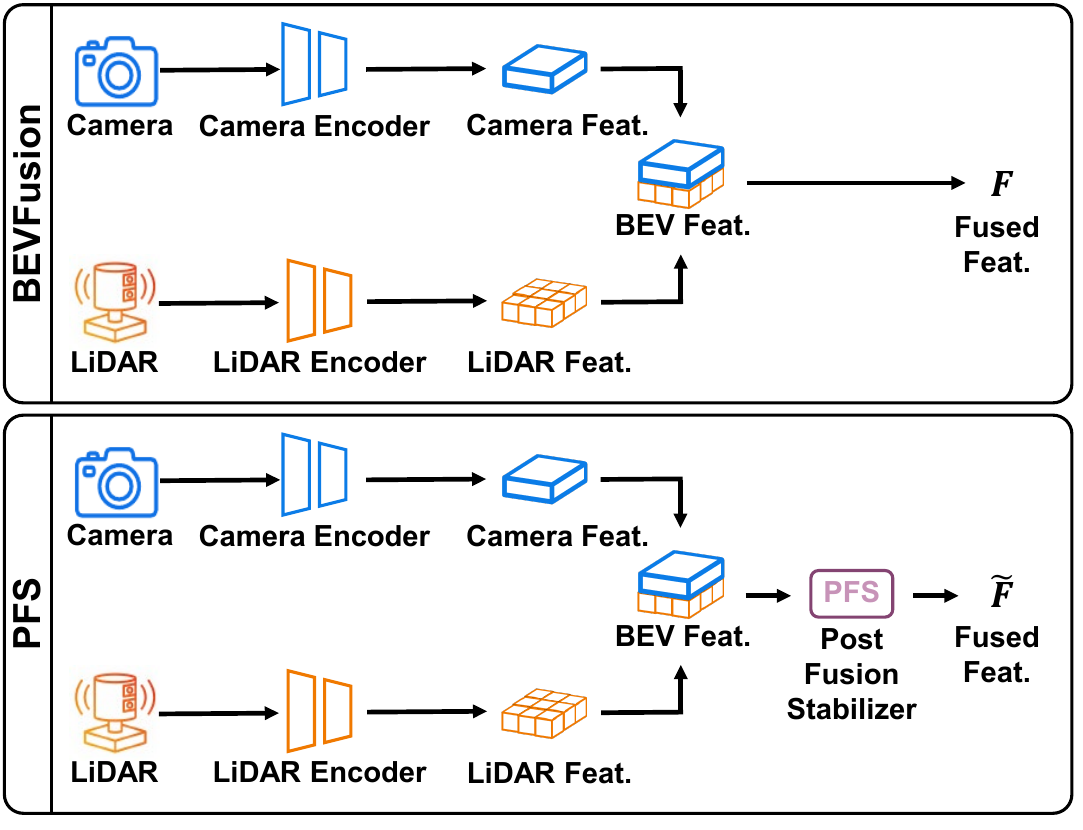}
    \caption{PFS inserts a lightweight correction module between the BEV fusion stage and the detection head of an existing detector.}
    \label{fig:pipeline}
    \vspace{-7mm}
\end{figure}

\begin{figure*}[t]
    \centering
    \includegraphics[width=0.8\textwidth, trim=0 3.9cm 5.1cm 0, clip]{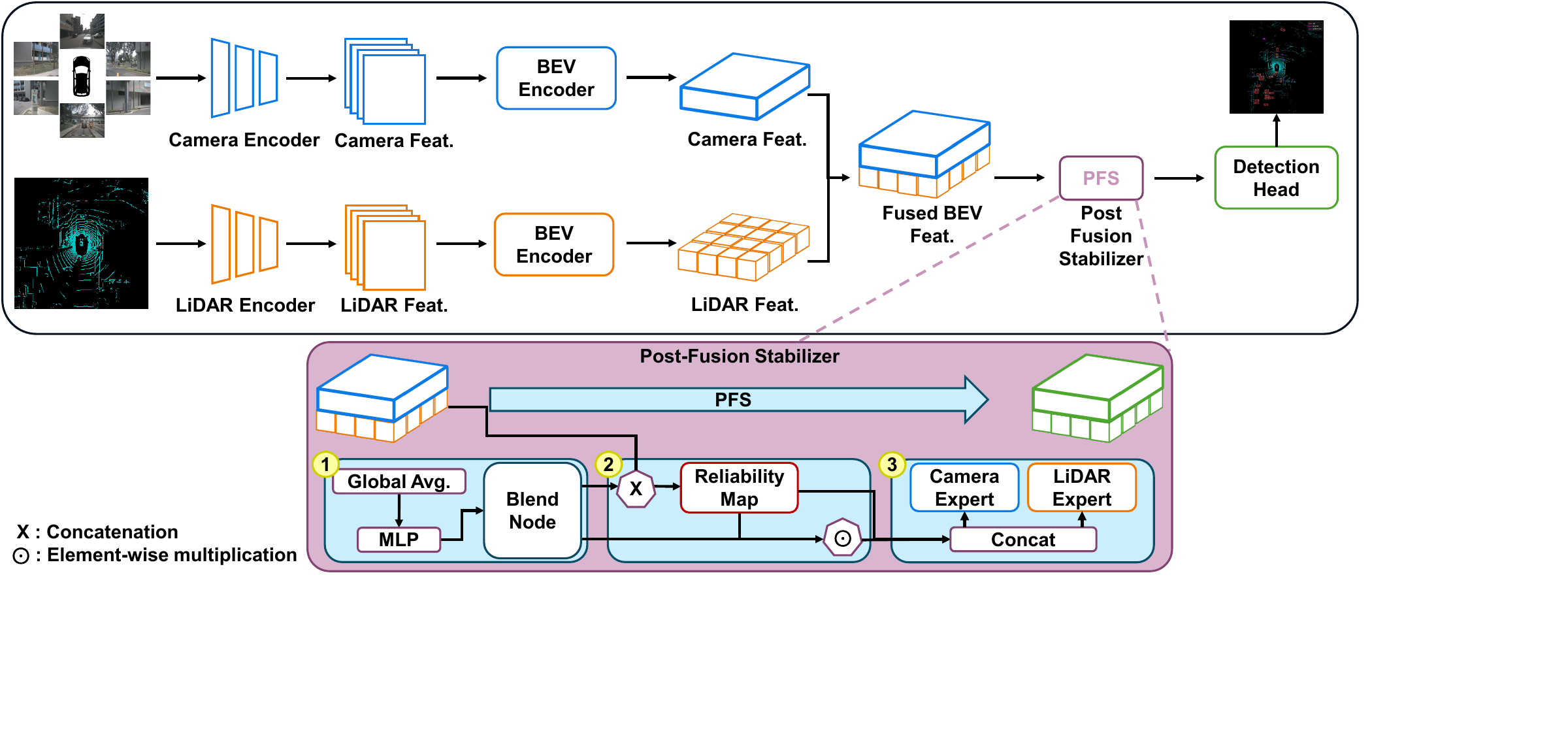}

    \caption{Overall PFS architecture. A lightweight Feature Stabilizer is placed between the fused BEV feature map and the frozen detection head. Operating on $\mathbf{F}\in\mathbb{R}^{C\times H\times W}$, it applies (1) channel shift normalization, (2) spatial reliability suppression, and (3) degradation aware gated residual correction with semantic and geometric experts to produce $\tilde{\mathbf{F}}$, which is then fed to the unchanged head. All stabilizer blocks are identity initialized for safe deployment.}
    \label{fig:architecture}
    \vspace{-6mm}
\end{figure*}

To address the integration overhead associated with backbone replacement, we instead focus on post-fusion refinement within the existing perception pipeline. We propose PFS, a lightweight module that attaches to the intermediate BEV representation of a fusion detector and refines it before the final detection head, as illustrated in Fig.~\ref{fig:pipeline}. This design preserves architectural structure while enabling robustness enhancement without modifying the core backbone or decoder. Our key observation is that many sensor corruptions produce structured distortions in the BEV feature space. Domain shifts induce distributional drift in feature statistics, partial sensor failures generate spatially localized degradations, and multi-modal corruption weakens discriminative cues. By operating directly in this shared spatial representation, PFS can correct these distortions in a modular and detector-agnostic manner. To our knowledge, prior work has not explored lightweight, identity initialized post fusion BEV correction as a mechanism for multi modal 3D detection.The contributions are as follows:

\begin{itemize}
    \item We introduce PFS, a robustness module for BEV-based camera-LiDAR fusion that refines intermediate BEV representations without modifying the backbone, fusion module, or detection head.
    \item We design PFS with three sequential correction blocks to address complementary BEV failure patterns: shift normalization for global drift, spatial reliability estimation for localized corruption, and gated residual correction for weakened sensing cues.
    \item We demonstrate that PFS achieved state-of-the-art (SOTA) performance among existing camera-LiDAR fusion methods under various sensor failure conditions and adverse weather scenarios.
\end{itemize}


\section{Background \& related works}

\subsection{BEV-Based Multi-Modal Fusion for 3D Detection}

The effectiveness of BEV as a unified fusion space was initially discussed by BEVFusion\cite{liu2024bevfusionmultitaskmultisensorfusion}, which demonstrated that preserving semantic density from cameras and geometric structure from LiDAR within a shared grid significantly outperforms single-modality pipelines. This paradigm has evolved into two primary branches: dense grid-level fusion  and sparse interaction-based fusion. Methods such as TransFusion\cite{zhou2024transfusionpredicttokendiffuse} and DeepInteraction\cite{yang2022deepinteraction3dobjectdetection} interleave cross-modal interaction layers to progressively exchange information, while UniTR\cite{wang2023unitrunifiedefficientmultimodal} and SparseFusion\cite{xie2023sparsefusionfusingmultimodalsparse} utilize transformers or instance-level candidates to mitigate the modality gap.

However, the shared BEV-based coupling of modalities introduces a significant structural vulnerability known as feature leakage. Because these methods are typically optimized for clean data, corrupted features from an unreliable sensor can \say{leak} into the shared BEV representation, distorting the final features consumed by the detection head. This phenomenon is particularly pronounced in robotic deployments where distribution shifts (e.g., weather) or partial sensor failures (e.g., LiDAR beam reduction) are common.

\subsection{Robustness Benchmarks and Resilient Fusion Strategies}
The vulnerability of fusion models to distribution shifts and hardware failures has prompted the development of specialized benchmarks, such as nuScenes-C\cite{dong2023benchmarking} and nuScenes-R\cite{yu2022benchmarkingrobustnesslidarcamerafusion}, which simulate realistic stressors like limited field-of-view, sensor occlusion, and total modality loss. These evaluation toolkits consistently reveal a common \say{geometric collapse} where standard architectures fail to maintain spatial reasoning once the primary LiDAR signal is disrupted.

In response, the research community has pursued three primary categories of resilient strategies. Architectural redesigns, exemplified by MetaBEV\cite{ge2023metabevsolvingsensorfailures} and UniBEV\cite{wang2024unibevmultimodal3dobject}, introduce modality-arbitrary decoders or advanced alignment rules to ensure the model remains functional even when specific sensor inputs are entirely missing. Alternatively, expert-based decoding strategies, such as MoME\cite{park2025resilientsensorfusionadverse}, utilize a mixture of experts specialized for different sensor combinations. By routing object queries based on local feature quality, these methods prevent harmful cross-modal entanglement and ensure that corrupted features do not degrade the final detection. Finally, targeted training approaches, including those used by CMT\cite{yan2023crossmodaltransformerfast} and UniTR\cite{wang2023unitrunifiedefficientmultimodal}, employ masked-modality training and stochastic dropout during the initial optimization phase. However, these training-based strategies still require modification of the optimization pipeline. This fosters modality independent representation resilient to transient sensor failures, and introduces a lightweight, identity initialized post fusion BEV correction for multimodal 3D detection.
\vspace{-2mm}

%
\section{Methodology}

We propose PFS, as shown in Fig. \ref{fig:architecture}, a lightweight module that attaches to the fused BEV feature map produced by a BEV-based fusion detector (e.g., BEVFusion~\cite{liu2024bevfusionmultitaskmultisensorfusion}). Let the host detector produce a fused feature tensor $\mathbf{F}_{\text{fused}} \in \mathbb{R}^{B \times C \times H \times W}$, where $B$ represents the batch size, $C$ denotes the feature channel dimension, and the spatial dimensions $H$ and $W$ correspond to the spatial height and width of the bird's-eye view grids. PFS intercepts $\mathbf{F}_{\text{fused}}$ and outputs a corrected tensor $\tilde{\mathbf{F}} \in \mathbb{R}^{B \times C \times H \times W}$, which is passed to the original detection head. Throughout training, the host detector remains fully frozen.

PFS consists of three sequential blocks designed as an identity-preserving residual stream. This architecture ensures that at initialization, the module is mathematically equivalent to an identity transform, preserving the baseline performance of the pretrained detector.

\subsection{Block 1: BEV Shift Normalization}
Camera domain shifts (e.g., \textit{low light}) induce global distributional drift in the BEV representation. Block 1 (~99K params) learns a conditional affine correction to stabilize these statistics. We compute a global context vector $\mathbf{g}$ via spatial average pooling and predict per-channel scale ($\boldsymbol{\gamma}$) and bias ($\boldsymbol{\beta}$) using a lightweight MLP (Linear 256$\rightarrow$128$\rightarrow$512) as
\begin{equation}
\mathbf{F}_{\text{shift}} = \sigma(\alpha) \cdot (\boldsymbol{\gamma} \odot \text{GN}(\mathbf{F}_{\text{fused}}) + \boldsymbol{\beta}) + (1 - \sigma(\alpha)) \cdot \mathbf{F}_{\text{fused}},
\end{equation} 
where $\sigma(\cdot)$ denotes the standard sigmoid function, which serves as a gating mechanism to control the influence of the normalization block. 

The term $\text{GN}(\cdot)$ represents Group Normalization, applied to maintain stable internal activations independently of the batch size $B$. The operator $\odot$ signifies channel-wise multiplication (Hadamard product) between the predicted scale $\boldsymbol{\gamma}$ and the normalized features. Crucially, $\alpha$ is a learnable scalar parameter initialized to $-5.0$, ensuring that $\sigma(\alpha) \approx 0.0067$ at the start of training; this effectively creates an identity mapping that preserves the host detector's original performance while allowing the module to gradually learn necessary corrections..

\subsection{Block 2: Spatial Reliability Estimation}
To address localized LiDAR degradations (e.g., \textit{beam reduction}), Block 2 (664k params) estimates a per-pixel reliability map $\mathbf{R} \in [0,1]^{B \times 1 \times H \times W}$. The map is predicted from the concatenation of $\mathbf{F}_{\text{shift}}$ and the raw LiDAR BEV features $\mathbf{F}_{\text{lidar}}$; if the latter is unavailable, the block degrades gracefully by duplicating $\mathbf{F}_{\text{shift}}$ as input as

\begin{equation}
\mathbf{R} = 
\begin{cases}
\sigma(\text{ConvNet}(\mathbf{F}_{\text{shift}} \oplus \mathbf{F}_{\text{lidar}})),  \text{if} \quad \exists \mathbf{F}_{\text{lidar}} = True \\
\sigma(\text{ConvNet}(\mathbf{F}_{\text{shift}} \oplus \mathbf{F}_{\text{shift}})),  \text{if} \quad \exists \mathbf{F}_{\text{lidar}} = False
\end{cases}
\end{equation}

In this equation, $\text{ConvNet}(\cdot)$ refers to a lightweight Convolutional Neural Network consisting of three sequential $3 \times 3$ convolutional layers with $2 \times 2$ stride, followed by a transposed convolution to upsample the output back to the original spatial resolution $H \times W$. 

The filtered feature map is defined as $\mathbf{F}_{\text{clean}} = \mathbf{R} \odot \mathbf{F}_{\text{shift}}$, where $\odot$ denotes the element-wise product. To prevent the reliability map from drifting toward a pessimistic uniform bias (e.g., $R \approx 0.5$) during training, we introduce an \textbf{Anchor Loss}. For clean samples, we supervise $\mathbf{R}$ toward an identity target of $1.0$ as

\begin{equation}
L_{\text{total}} = L_{\text{det}} + \lambda_{\text{rel}}\,L_{\text{rel}}.
\end{equation}
where $L_{\text{det}}$ represents the frozen host's detection loss, comprising heatmap and regression terms, and $\lambda_{\text{rel}}$ is a hyperparameter balancing the reliability correction against the primary task. The auxiliary reliability loss $L_{\text{rel}}$ is formulated as

\begin{equation}
L_{\text{rel}}=
\mathbb{1}_{\text{corr}} \cdot \mathrm{BCE}(\mathbf{R}, \mathbf{T})
+ \alpha_{\text{anchor}} \cdot \mathbb{1}_{\text{clean}} \cdot \mathrm{BCE}(\mathbf{R}, 1.0),
\end{equation}
In this equation, $\mathbb{1}_{\text{corr}}$ and $\mathbb{1}_{\text{clean}}$ are binary indicator variables while the term $\mathrm{BCE}(\cdot)$ denotes the Binary Cross Entropy function. The reliability target $\mathbf{T}$ is a per-BEV-cell point-density ratio defined as
\begin{equation}
T_{ij} = \text{clamp}\left(\frac{D^{\text{corr}}{ij}}{D^{\text{clean}}{ij}},\ 0,\ 1\right),
\end{equation}
where $D_{ij}$ is the point count in BEV cell $(i,j)$ after voxelization onto a 108$\times$108 grid. Semantically, $T_{ij} = 1$ indicates a fully preserved cell, while $T_{ij} = 0$ corresponds to total point removal. For edge cases where $D^{\text{clean}}_{ij} = 0$, we set $T_{ij} = 1$ to ensure that only regions with pre-existing point data contribute to the gradient.

\subsection{Block 3: Expert Correction and Inpainting}
Block 3 (2.5M params) recovers lost information in regions suppressed by Block 2 (e.g., camera \textit{dropout}). Unlike previous blocks, Block 3 utilizes the reliability map $\mathbf{R}$ as a \say{hole map} to guide its experts. To identify missing cues, the semantic ($E_s$) and geometric ($E_g$) experts receive the concatenation of $\mathbf{F}_{\text{clean}}$ and $\mathbf{R}$, defined as
\begin{equation}
\Delta\mathbf{F} = w_s \cdot E_s([\mathbf{F}_{\text{clean}}; \mathbf{R}]) + w_g \cdot E_g([\mathbf{F}_{\text{clean}}; \mathbf{R}]).
\end{equation}
This conditioning allows experts to specialize in \say{inpainting} regions where $\mathbf{R} \approx 0$. A spatial gate $\mathbf{G}$ is implemented to control the correction strength. By feeding $\mathbf{R}$ directly into the gate, the module learns to open only in unreliable regions:
\begin{equation}
\mathbf{G} = \sigma(\text{Conv}([\mathbf{F}_{\text{clean}}; \mathbf{R}])).
\end{equation}
The final corrected output $\tilde{\mathbf{F}}$ is a residual added to $\mathbf{F}_{\text{shift}}$, effectively bypassing Block 2's suppression to restore signal from the normalized stream:
\begin{equation}
\tilde{\mathbf{F}} = \mathbf{F}_{\text{shift}} + \mathbf{G} \odot \Delta\mathbf{F}.
\end{equation}
To ensure safe deployment, Block 3 is initialized as an identity transform. While the experts $E_s$ and $E_g$ use standard Kaiming initialization, the final gate $\mathbf{G}$ is initialized with a bias of $-4.0$. This ensures $\sigma(-4.0) \approx 0.018$ at the start of training, effectively disabling the experts and preserving the normalized BEV features $\mathbf{F}_{\text{shift}}$ until the gate is driven open by the detection loss. Similarly, Block 2 is initialized with a bias of $+4.0$ so that the reliability map $\mathbf{R} \approx 1.0$ initially, preventing any premature feature suppression.



\begin{figure*}[t]
    \centering
    \includegraphics[width=\textwidth, trim=1.6cm 6cm 1.6cm 1cm, clip]{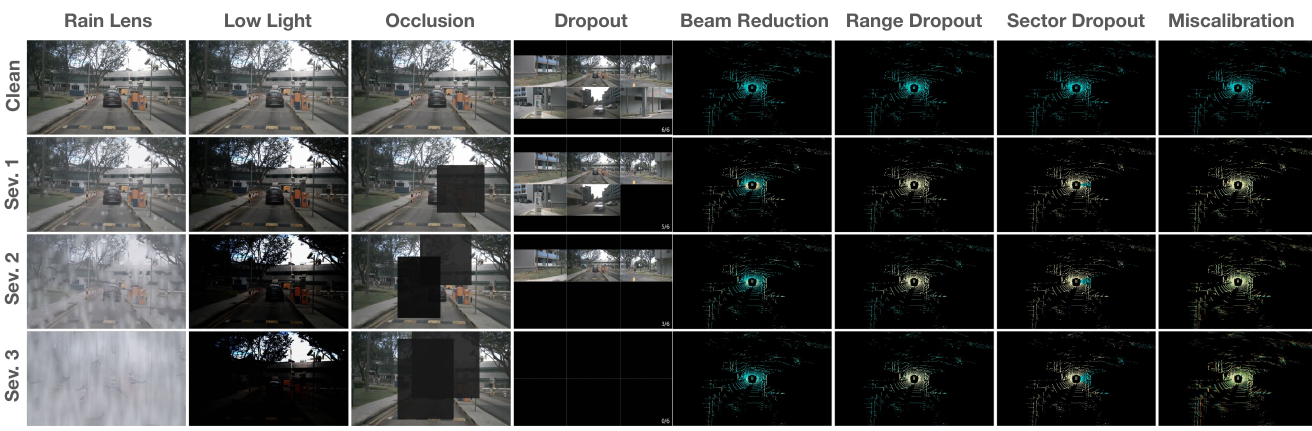}
    \caption{LiDAR and Camera corruptions at each severity levels.}
    \label{fig:cam_lidar_corrupt}
    \vspace{-4mm}
\end{figure*}

\subsection{Staged Curriculum Training}
\label{sec:training_details}


\begin{table}[ht]
\centering
\footnotesize
\normalsize
\setlength{\tabcolsep}{5pt}
\renewcommand{\arraystretch}{1.1}

\resizebox{\columnwidth}{!}{
\begin{tabular}{@{}l|l|c|c|c@{}}
\toprule
\textbf{Sensor} & \textbf{Corruption Type} & \textbf{L1} & \textbf{L2} & \textbf{L3} \\ \midrule
\multirow{4}{*}{Camera} 
& Rain Lens ($\alpha$ / drops) & 0.15 / 80 & 0.45 / 200 & 0.70 / 350 \\
\cmidrule(lr){2-5}
& Low Light ($\gamma$ / $\sigma$) & 1.8 / 0.018 & 3.2 / 0.045 & 5.5 / 0.10 \\
\cmidrule(lr){2-5}
& Occlusion (Area $\mathcal{C}$) & 15\% & 35\% & 60\% \\
\cmidrule(lr){2-5}
& Dropout (Views $N$) & 1 & 3 & 6 \\ \midrule
\multirow{4}{*}{LiDAR} 
& Beam Red. (Step $k$) & 2 (16-ch) & 4 (8-ch) & 8 (4-ch) \\
\cmidrule(lr){2-5}
& Range Drop ($P_{max} @ r_{max}$) & 0.70 @ 50m & 0.85 @ 45m & 0.95 @ 40m \\
\cmidrule(lr){2-5}
& Sector Drop (Azim. $\theta$) & 30$^\circ$ & 60$^\circ$ & 90$^\circ$ \\
\cmidrule(lr){2-5}
& Miscalib. ($\Delta\theta, \Delta t$) & 0.23$^\circ, 0.04$m & 0.85$^\circ, 0.08$m & 2.0$^\circ, 0.20$m \\ \bottomrule
\end{tabular}
}
\caption{\textbf{Sensor Corruption Severity Levels.} Parameters define the physical intensity of simulated degradation across three severity levels.}
\label{tab:corruption_params}
\vspace{-4mm}
\end{table}

To train PFS, we freeze the host detector and optimize only the stabilizer parameters in a three-stage curriculum. This phased approach prevents catastrophic feature drift and encourages block specialization. Throughout training, we apply a stochastic mix of sensor corruptions where camera and LiDAR perturbations are drawn independently with probability $p=0.30$. This results in a training distribution of 49\% clean, 21\% camera-only, 21\% LiDAR-only, and 9\% simultaneous corruptions, ensuring the model maintains clean-domain performance while seeing sufficient failure cases. For corruptions details deferred to Sec.~\ref{subsubsec:corruptions}.

\textbf{Stage 1: Shift Normalization.}
We optimize Block 1 for 6 epochs using a limited pool of global corruptions (Rain Lens, Low Light, Beam Reduction, and Miscalibration). We use AdamW with an initial learning rate of $2\times 10^{-4}$ and cosine annealing.

\textbf{Stage 2: Reliability Calibration.}
We jointly train Blocks 1 and 2 for 4 epochs using the full eight-corruption suite. We set $\lambda_{\text{rel}} = 1.0$ and introduce the Anchor Loss weighted at $\alpha_{\text{anchor}} = 0.2$. This anchor acts as a counter-force against downward bias drift during the 70\% of samples that lack LiDAR corruption, preventing the reliability map from uniformly collapsing toward zero. The learning rate is reduced to $1\times 10^{-4}$ to stabilize refinement.

\textbf{Stage 3: Expert Correction.}
We add Block 3 (~2.5M params) and optimize it for 4 epochs while freezing Blocks 1 and 2. This freezing strategy preserves the calibrated reliability map developed in Stage 2, ensuring Block 3 receives a stable input distribution and a fixed \say{hole map} for reconstruction. This prevents the gradient starvation and clean-domain regression observed in unconstrained joint training. We use a higher learning rate of $2\times 10^{-4}$ in this stage to overcome the initial gradient vanishing caused by the small identity-initialized gate.

\textbf{Losses.}
All stages utilize the standard focal and $L_1$ detection losses from the frozen host head. Training is performed with a batch size of 4 and gradient clipping (max norm = 5.0, except Stage 1) to ensure convergence. Aside from the point-density supervision in Stage 2, the module requires no additional annotations

\section{Experiments}
\label{section:Experiments_and_results}

We evaluate PFS in two settings. First, we run controlled simulation experiments on NuScenes\cite{nuscenes2019} \textit{val} and NuScenes-C\cite{dong2023benchmarking} to quantify robustness under a fixed corruption suite. Second, we evaluate on a real world dataset collected by our lab vehicle to validate transfer beyond synthetic perturbations.

\subsection{Simulation Experiments}
\label{subsec:sim_nuscenes}

\subsubsection{Experimental Setup}
\textbf{Dataset.} We evaluate on the nuScenes dataset\cite{nuscenes2019} and nuScenes-C\cite{dong2023benchmarking}, using the official train and validation splits with 10 object classes.

\textbf{Metrics.} Following the standard nuScenes evaluation protocol, we report mean Average Precision (mAP) and the nuScenes Detection Score (NDS). The mAP is calculated in the BEV across ten distinct object classes, utilizing center distance thresholds of 0.5m, 1m, 2m, and 4m. To provide a comprehensive performance metric, the NDS is derived by aggregating the mAP with mean errors related to translation, scale, orientation, velocity, and attributes.


\textbf{Baselines.} We evaluate PFS as a post fusion module on top of BEV based camera and LiDAR fusion detectors implemented in MMDetection3D \cite{mmdet3d2020}.

\subsubsection{Corruption suite and evaluation design}


\begin{table*}[t]
\centering
\normalsize
\renewcommand{\arraystretch}{1.5}
\resizebox{\textwidth}{!}{
\begin{tabular}{@{}l|cc|cc|cc|cc|cc|cc|cc|cc|cc@{}}
\toprule
\multicolumn{1}{c|}{} 
& \multicolumn{2}{c|}{} 
& \multicolumn{8}{c|}{\textbf{LiDAR corruptions}} 
& \multicolumn{8}{c}{\textbf{Camera corruptions}} \\
\cline{4-11} \cline{12-19}

\multirow{2}[-1]{*}{\textbf{Method}} 
& \multicolumn{2}{c|}{\textbf{Clean}} 
& \multicolumn{2}{c|}{\textbf{Beam Reduction}} 
& \multicolumn{2}{c|}{\textbf{Range Dropout}} 
& \multicolumn{2}{c|}{\textbf{Sector Dropout}} 
& \multicolumn{2}{c|}{\textbf{Miscalibration}} 
& \multicolumn{2}{c|}{\textbf{Rain Lens}} 
& \multicolumn{2}{c|}{\textbf{Low Light}} 
& \multicolumn{2}{c|}{\textbf{Occlusion}} 
& \multicolumn{2}{c}{\textbf{Dropout}} \\

\multicolumn{1}{c|}{} 
& \multicolumn{2}{c|}{} 
& \multicolumn{2}{c|}{\textit{8-beams}} 
& \multicolumn{2}{c|}{\textit{95\% @ 40m}} 
& \multicolumn{2}{c|}{\textit{90$^\circ$ Sector}} 
& \multicolumn{2}{c|}{\textit{2.0$^\circ$, 0.2m}} 
& \multicolumn{2}{c|}{\textit{350 drops}} 
& \multicolumn{2}{c|}{\textit{$\gamma=5.5$}} 
& \multicolumn{2}{c|}{\textit{60\% Area}} 
& \multicolumn{2}{c}{\textit{6 drops}}\\

\cline{2-19}
& mAP & NDS 
& mAP & NDS 
& mAP & NDS 
& mAP & NDS 
& mAP & NDS 
& mAP & NDS 
& mAP & NDS 
& mAP & NDS 
& mAP & NDS \\
\midrule

BEVFusion $^{\dagger}$~\cite{liu2024bevfusionmultitaskmultisensorfusion} 
& 69.2 & 59.7
& 12.1 & 24.8
& 52.6 & 50.0
& 54.7 & 52.4
& 46.8 & 44.1
& 60.0 & 55.0
& 58.8 & 54.4
& 64.6 & 57.3
& 62.6 & 56.3 \\

UniBEV$^{\dagger}$~\cite{wang2024unibevmultimodal3dobject} 
& 65.3 & 57.2
& 25.6 & 29.9
& 53.2 & 50.0
& 56.5 & 52.2
& 48.0 & 45.1
& 56.5 & 52.8
& 55.4 & 52.3
& 61.5 & 55.5
& 56.7 & 53.1 \\

CMT$^{\dagger}$~\cite{yan2023crossmodaltransformerfast} 
& 70.1 & 72.9
& \underline{32.6} & 45.7
& \underline{58.5} & 65.2
& 60.7 & 67.0
& 50.1 & 58.0
& 58.9 & 66.6
& 55.4 & 64.3
& 64.1 & 70.2
& 62.0 & 68.6 \\

DeepInteraction$^{\dagger}$~\cite{yang2022deepinteraction3dobjectdetection} 
& 68.9 & 58.3
& 7.74  & 22.6
& 49.9 & 47.84
& 53.6 & 50.2
& 45.2 & 42.0
& 54.4 & 49.1
& 58.8 & 51.6
& 59.5 & 52.7
& 53.7 & 47.3 \\

SparseFusion$^{\dagger}$~\cite{xie2023sparsefusionfusingmultimodalsparse} 
& \underline{70.5} & 72.5
& 5.2 & 23.3
& \textbf{68.2} & 71.2
& 32.4 & 50.8
& 47.2 & 56.1
& \underline{60.4} & 66.9
& 55.9 & 64.4
& 64.4 & 70.7
& 55.4 & 63.8 \\

MoME$^{\dagger}$~\cite{park2025resilientsensorfusionadverse} 
& \textbf{72.1} & 74.4
& 23.1 & 41.4
& 60.0 & 66.8
& \textbf{61.2} & 67,6
& \underline{52.6} & 60.2
& 50.7 & 62.7
& 44.6 & 59.5
& 64.1 & 69.7
& \underline{63.8} & 69.8 \\

\midrule
BEVFusion~\cite{liu2024bevfusionmultitaskmultisensorfusion}\textbf{ + PFS} 
& 69.5 & 71.2
& 17.5 & 27.1
& 53.4 & 60.6
& 56.9 & 63.2
& 47.5 & 55.5
& \textbf{63.6} & 68.3
& \textbf{63.2} & 64.9
& \textbf{65.3} & 69.0
& \textbf{63.8} & 68.7 \\

UniBEV~\cite{wang2024unibevmultimodal3dobject}\textbf{ + PFS}
& 65.3 & 57.2
& \textbf{33.8} & 35.5
& 53.4 & 50.1
& \underline{60.7} & 66.5
& \textbf{52.9} & 59.4
& 58.1 & 62.8
& \underline{58.9} & 60.9
& \underline{64.7} & 63.6
& 58.4 & 64.0 \\

\bottomrule
\end{tabular}
}
\caption{\textbf{Comparison under various sensor corruptions and sensor failures on nuScenes \textit{val} dataset.}’†’ indicates reproduced results using their open-source code. \textit{Italic} denotes the degree of sensor failure. All sensor corruptions severity are level 3. PFS achieves state-of-the-art performance across most tasks (\textbf{BEST}/ \underline{SECOND BEST}).}
\label{table_1}
\vspace{-5mm}
\end{table*}
\begin{table}[!t]
  \centering
  \footnotesize 
  
  \begin{tabular}{@{}ccccc@{}}
    \toprule
    \multicolumn{1}{c}{Method} & \multicolumn{1}{c}{Modality} & Fog & Snow & Sunlight \\
    \midrule
    PGD \cite{PGD} & C & 12.8 & 2.3 & 22.8 \\
    FCOS3D \cite{fcos3d} & C & 13.5 & 2.0 & 17.2 \\
    DETR3D \cite{detr3d} & C & 27.9 & 5.1 & 41.6 \\
    BEVFormer \cite{li2022bevformerlearningbirdseyeviewrepresentation} & C & 32.8 & 5.7 & 41.7 \\
    PointPillars \cite{pointpillar} & L & 24.5 & 27.6 & 23.7 \\
    SSN \cite{ssn} & L & 41.6 & 46.4 & 40.3 \\
    CenterPoint \cite{centerpoint} & L & 43.8 & 55.9 & 54.2 \\
    FUTR3D \cite{futr3d} & LC & 53.2 & 52.7 & 57.7 \\
    Transfusion \cite{zhou2024transfusionpredicttokendiffuse} & LC & 53.7 & 63.3 & 55.1 \\
    BEVFusion \cite{liu2024bevfusionmultitaskmultisensorfusion} & LC & 54.1 & 62.8 & 64.4 \\
    DeepInteraction \cite{yang2022deepinteraction3dobjectdetection} & LC & 54.8 & 62.4 & 64.9 \\
    CMT \cite{yan2023crossmodaltransformerfast} & LC & 66.3 & 62.6 & 63.6 \\
    MoME \cite{park2025resilientsensorfusionadverse} & LC & \underline{67.9} & 63.5 & \underline{65.2} \\
    \midrule
    BEVFusion~\cite{liu2024bevfusionmultitaskmultisensorfusion}\textbf{ + PFS} 
    & LC & \textbf{67.9} & \textbf{65.5} & \textbf{65.6} \\
    
    UniBEV~\cite{wang2024unibevmultimodal3dobject}\textbf{ + PFS}
    & LC & 65.7 & \underline{63.6} & 61.3 \\

    \bottomrule
    \end{tabular}
  \caption{\textbf{Comparison under extreme weather conditions on nuScenes-C.} `L' and `C' represent camera, LiDAR, respectively (\textbf{BEST}/ \underline{SECOND BEST}). }
  \vspace{-5mm}
  \label{tab_2}
\end{table}


\label{subsubsec:corruptions}
Our evaluation utilizes a focused suite of eight corruptions spanning camera and LiDAR modalities across three severity levels, as shown in table \ref{tab:corruption_params}. We select perturbations that specifically target the failure modes PFS is designed to stabilize: global distribution shift, spatially localized voids, and cross-sensor misalignment.

\textit{\textbf{Camera corruptions (Fig.~\ref{fig:cam_lidar_corrupt}).}}
\textit{Rain Lens} models water accumulation via global haze $\alpha$, quadratic-falloff droplets, and vertical smears. \textit{Low Light} simulates photon-limited environments using power-law darkening $I_{\text{dark}} = I^{\gamma}$ and signal-dependent shot noise scaled by $\sqrt{I_{\text{dark}} + 10^{-4}}$. \textit{Occlusion} applies opaque center-biased masks to obstruct the horizon, while \textit{Dropout} simulates transmission failure by zeroing $N$ camera views.

\textit{\textbf{LiDAR corruptions (Fig.~\ref{fig:cam_lidar_corrupt}).}}
\textit{Beam Reduction} reduces vertical resolution by retaining every $k$-th elevation bin. \textit{Range Dropout} simulates atmospheric attenuation via a piecewise-linear drop probability $P(r)$ reaching $P_{max}$ at range $r_{max}$. \textit{Sector Dropout} removes points within an azimuth sector $\theta$ to model \say{blind spots} and \textit{Miscalibration} introduces rigid extrinsic perturbations $(\Delta\theta, \Delta t)$ to challenge spatial alignment.

For camera only tests, we corrupt all six surround view images while keeping LiDAR clean. For LiDAR only tests, we corrupt the LiDAR point cloud while keeping camera images clean. To ensure fair comparison across methods, all corruptions are generated deterministically using a fixed per sample seed, so every method is evaluated on the same corrupted samples.

\subsubsection{Main results on NuScenes \textit{Val} and NuScenes-C}

The performance of the proposed PFS module is evaluated across two benchmarks: a focused suite of level-3 sensor corruptions (Table~\ref{table_1}) and the simulated failure scenarios of the nuScenes-C dataset (Table~\ref{tab_2}). These results quantify the effectiveness of PFS as a lightweight stabilizer attachment for existing BEV-based fusion detectors compared to their unmodified counterparts.


\begin{figure*}[t]
    \centering
    \includegraphics[width=\textwidth, trim=5cm 10cm 5cm 0cm, clip]{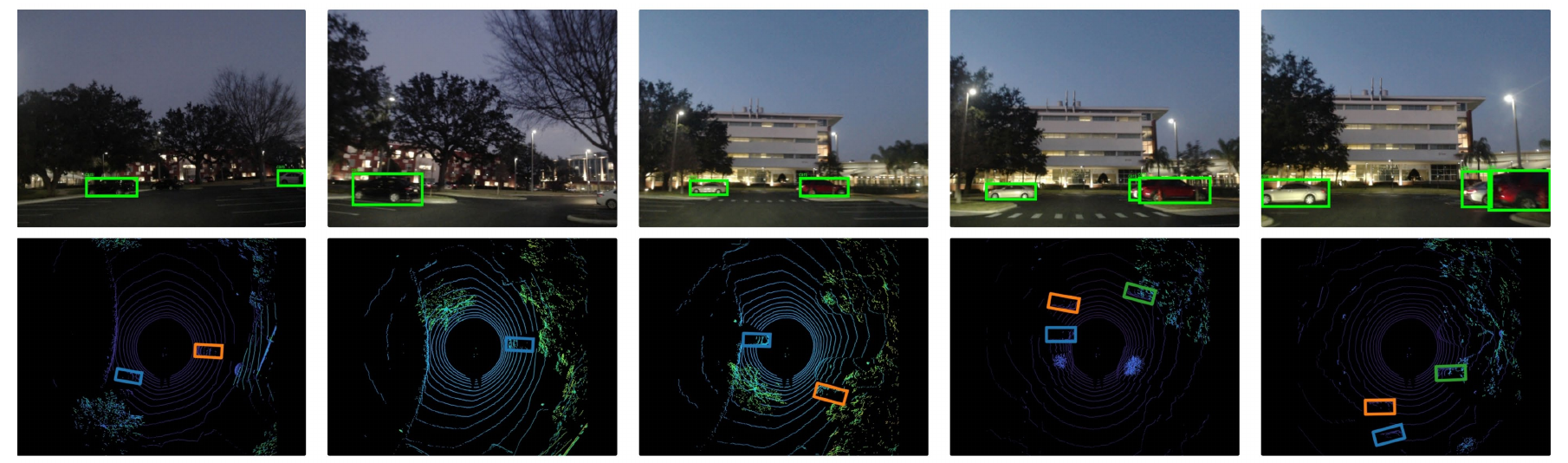}
    \caption{Real time camera and LiDAR synchronization during deployment. Each card shows a timestamp aligned RGB frame (top) and raw LiDAR point cloud (bottom) in the ego frame.}
    \label{fig:realtime_sync}
\end{figure*}

\textbf{Robustness under NuScenes \textit{Val}.}
Table~\ref{table_1} summarizes the performance of PFS when integrated with the BEVFusion backbone under Level 3 environmental and sensor perturbations. Unlike previous adaptive methods that often suffer from regression on clean data, PFS preserves and slightly improves the baseline performance of BEVFusion by achieving 69.5 mAP and 71.2 NDS. This validates the efficacy of our Identity-Initialization and Anchor Loss strategies, which were specifically designed to prevent the reliability map from drifting toward a sub-optimal uniform bias during training.

When attached on BEVFusion~\cite{liu2024bevfusionmultitaskmultisensorfusion} PFS achieves SOTA results across all camera failure modes, reaching 63.8 mAP in the extreme 6-camera Dropout scenario where it matched specialized resilient architectures like MoME and CMT. It outperforms the benchmark in Rain Lens and Low Light conditions, delivering gains of +3.6\% and +4.4\% mAP over the baseline respectively. While absolute LiDAR performance remains constrained by the host's geometric extraction capabilities, PFS still significantly enhances baseline resilience. In the severe 8-beam reduction case, PFS improves the BEVFusion baseline by +5.4 points in mAP, demonstrating robust stabilization of sparse geometric features.

The integration of PFS with UniBEV~\cite{wang2024unibevmultimodal3dobject} further demonstrates the module's ability to overcome architectural ceilings in geometric reasoning. UniBEV + PFS achieves the overall best performance in LiDAR-centric corruptions, setting new benchmarks in Beam Reduction (33.8 mAP) and Miscalibration (52.9 mAP). This result confirms that PFS acts as a \say{Stability Multiplier}; by stabilizing the underlying geometric features of a more robust backbone like UniBEV, the system achieves a higher absolute performance ceiling than BEVFusion + PFS, particularly in cases of extreme sparsity and spatial misalignment. In the Sector Dropout case, UniBEV + PFS achieves 60.7 mAP, nearly matching the specialized MoME\cite{park2025resilientsensorfusionadverse} architecture while utilizing a significantly lighter modular plug-in.

\textbf{Robustness under NuScenes-C.}
The resilience of PFS against extreme environmental conditions is evaluated using the nuScenes-C benchmark, focusing on Fog, Snow, and Sunlight scenarios (Table~\ref{tab_2}). Integration with the BEVFusion backbone yields state-of-the-art results across all evaluated weather types, matching or exceeding specialized resilient architectures like MoME and CMT.Specifically, BEVFusion + PFS achieves a +25.5\% relative mAP improvement in Fog compared to its baseline, reaching 67.9 mAP. In Snow and Sunlight conditions, the module provides steady absolute gains of +4.3\% and +1.9\% mAP respectively. These improvements confirm that the Shift Normalization in Block 1 effectively calibrates for atmospheric-induced distribution shifts that typically degrade the camera-heavy semantic features consumed by the fusion head.The UniBEV + PFS configuration also demonstrates strong performance, particularly in Snow where it achieves 63.6 mAP and secures the second-best result in the category.


\subsubsection{Block ablation}
We ablate the three sequential blocks of the proposed PFS module by progressively enabling them on BEVFusion~\cite{liu2024bevfusionmultitaskmultisensorfusion} and UniBEV~\cite{wang2024unibevmultimodal3dobject}. As summarized in Table~\ref{tab:ablation}, the results support our design rationale: Block 1 corrects global feature drift, Block 2 suppresses localized unreliable regions, and Block 3 leverages the reliability map to apply targeted expert correction. These mechanisms are complementary, and their combination yields the strongest robustness.

\textbf{BEVFusion.} The baseline is highly sensitive to LiDAR failures, most notably Beam Reduction at 12.1 mAP. Enabling Block 1 provides the first consistent gain in camera corruptions by stabilizing global statistics, improving Low Light from 58.8 to 61.4 mAP with near clean parity (69.2 to 68.7 mAP). Adding Block 2 introduces spatially aware suppression that primarily benefits geometric corruptions, raising Beam Reduction to 14.9 mAP, a +23.1\% relative gain over the baseline. The full pipeline further improves both camera and LiDAR failures, with Rain Lens reaching 63.6 mAP and Beam Reduction reaching 17.5 mAP. Importantly, Block 3 also recovers and slightly improves clean accuracy (68.9 to 69.5 mAP) while producing the largest NDS gains across settings, indicating more stable attribute and motion predictions under corruption.

\textbf{UniBEV.} UniBEV starts more resilient to LiDAR sparsity than BEVFusion but still degrades under extreme failures (Beam Reduction 25.6 mAP, Miscalibration 48.0 mAP). Block 1 mainly strengthens camera robustness, improving Low Light from 55.4 to 57.6 mAP without affecting clean metrics. Block 2 yields a clear NDS jump under Rain Lens (52.8 to 56.5), consistent with suppressing droplet induced localized noise before it propagates to the head. The full pipeline shows the strongest synergy under severe geometric failures, improving Beam Reduction to 33.8 mAP, a +32.0\% relative gain, and correcting spatial offsets in Miscalibration (48.0 to 52.9 mAP). The corresponding NDS increase under Miscalibration is especially large (45.1 to 59.4, +31.7\%), confirming that gated expert correction substantially stabilizes downstream prediction quality even when sensor alignment is compromised.

\begin{table}[!t]
\centering
\footnotesize
\setlength{\tabcolsep}{2.6pt}
\renewcommand{\arraystretch}{1.05}

\resizebox{\columnwidth}{!}{
\begin{tabular}{@{}l |cc cc cc cc cc@{}}
\toprule
\multirow{2}{*}{\textbf{Method}} 
& \multicolumn{2}{c}{\textbf{Clean}} 
& \multicolumn{2}{c}{\textbf{Rain Lens}} 
& \multicolumn{2}{c}{\textbf{Low Light}} 
& \multicolumn{2}{c}{\textbf{Beam Red.}} 
& \multicolumn{2}{c}{\textbf{Miscalib.}} \\
\cmidrule(lr){2-3}\cmidrule(lr){4-5}\cmidrule(lr){6-7}\cmidrule(lr){8-9}\cmidrule(lr){10-11}
& mAP & NDS & mAP & NDS & mAP & NDS & mAP & NDS & mAP & NDS \\
\midrule

A0: \textbf{BEVFusion}\cite{liu2024bevfusionmultitaskmultisensorfusion}    
& 69.2 & 59.7
& 60.0 & 55.0
& 58.8 & 54.4
& 12.1 & 24.8
& 46.8 & 44.1 \\
A1: + Block 1        
& 68.7 & 60.0
& 61.6 & 56.0
& 61.4 & 56.7
& 11.9 & 25.5 
& 46.8 & 44.6 \\
A2: + Blocks 1 and 2 
& 68.9 & 60.0 
& 62.3 & 56.5
& 61.8 & 57.0
& 14.9 & 25.0 
& 47.0 & 44.6 \\
A3: + All blocks     
& \textbf{69.5} & \textbf{71.2}
& \textbf{63.6} & \textbf{68.3}
& \textbf{63.2} & \textbf{64.9}
& \textbf{17.5} & \textbf{27.1}
& \textbf{47.5} & \textbf{55.5} \\
\midrule

A0: \textbf{UniBEV}\cite{wang2024unibevmultimodal3dobject}    
& 65.3 & 57.2
& 56.5 & 52.8 
& 55.4 & 52.3
& 25.6 & 29.9
& 48.0 & 45.1 \\
A1: + Block 1        
& 65.3 & 57.2
& 56.9 & 53.2
& 57.6 & 53.4
& 26.4 & 30.0
& 48.2 & 45.1 \\
A2: + Blocks 1 and 2 
& 65.2 & 57.2 
& 57.1 & 56.5 
& 57.6 & 56.8
& 26.5 & 30.0
& 48.2 & 45.1 \\
A3: + All blocks     
& \textbf{65.3} & \textbf{57.2}
& \textbf{58.1} & \textbf{62.8} 
& \textbf{58.9} & \textbf{60.9}
& \textbf{33.8} & \textbf{35.5}
& \textbf{52.9} & \textbf{59.4} \\

\bottomrule
\end{tabular}%
}

\caption{\textbf{Block ablation across detectors on nuScenes val.} Each detector is evaluated with progressively enabled blocks: A0 (base), A1 (+Block 1), A2 (+Blocks 1\&2), A3 (+All blocks).}
\label{tab:ablation}
\end{table}

\begin{figure}[ht]
    \centering
    \includegraphics[width=0.8\columnwidth, trim=0cm 0.5cm 0cm 0.5cm, clip]{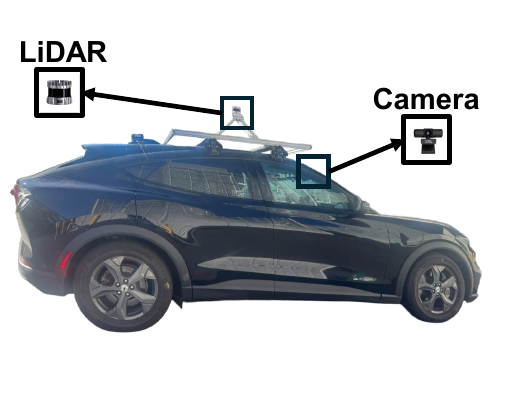}
    \caption{Real-world data collection and deployment platform}
    \label{fig:demo_vehicle}
\end{figure}

\subsection{Physical evaluation on real world dataset}
\label{subsec:physical}

\subsubsection{Deployment and Evaluation Protocol}

Fig. \ref{fig:realtime_sync} illustrates the synchronized camera–LiDAR pairs are processed using the same frozen host detectors evaluated in simulation, with PFS attached as a post-fusion head operating on the fused BEV feature tensor prior to the detection head. No additional fine-tuning or calibration adjustment is performed on the collected physical dataset.

The timestamp-aligned pairs demonstrate stable cross-modal synchronization and consistent spatial alignment under both daytime and low-illumination conditions. These real-world sequences validate integration compatibility with physical sensing hardware and confirm that PFS operates within a standard sensor pipeline without modification to upstream encoders or fusion backbones.

\subsubsection{Platform and data collection}
We deploy the PFS augmented detector on a lab vehicle equipped with a forward-facing RGB camera and a roof mounted 32-beam LiDAR, as shown in Fig.~\ref{fig:demo_vehicle}. Data is recorded at 10 Hz for both modalities using timestamp-based logging.

Camera-LiDAR synchronization is performed using a nearest-timestamp pairing strategy with a maximum tolerance of 50 ms. For each LiDAR sweep, the temporally closest RGB frame is selected, and frame pairs exceeding the tolerance threshold are discarded to ensure alignment consistency. All synchronized pairs are represented in the ego coordinate frame using fixed intrinsic and extrinsic calibration parameters. We collect urban driving sequences under clear daytime and low-illumination evening conditions. From these sequences, 200 synchronized frames are prepared for evaluation to assess transfer beyond synthetic corruption settings.

\subsubsection{Results on Physical Dataset}

\begin{table}[!t]
\centering
\footnotesize
\setlength{\tabcolsep}{3.5pt}
\renewcommand{\arraystretch}{1.2}
\resizebox{\columnwidth}{!}{%
\begin{tabular}{@{}l|cc|cccc@{}}
\toprule
\multirow{2}{*}{\textbf{Method}} & \multicolumn{2}{c|}{\textbf{mAP}} & \multicolumn{4}{c}{\textbf{Efficiency / Overhead}} \\
\cmidrule(lr){2-3} \cmidrule(lr){4-7}
& \textbf{Day} & \textbf{Night} & \textbf{Params (M)} & \textbf{Latency (ms)} & \textbf{FPS} & \textbf{Overhead (\%)} \\
\midrule
BEVFusion~\cite{liu2024bevfusionmultitaskmultisensorfusion} 
& 32.87 & 24.08 & 40.8 & 70.9 & 14.1 & - \\
+ \textbf{PFS} (Proposed) 
& \textbf{35.33} & \textbf{29.20} & 44.1 & 76.6 & 13.1 & 8.1\% \\
\midrule
UniBEV~\cite{wang2024unibevmultimodal3dobject} 
& 33.87 & 24.75 & 72.7 & 417.0 & 2.4 & - \\
+ \textbf{PFS} (Proposed) 
& \textbf{36.65} & \textbf{28.83} & 76.0 & 436.0 & 2.3 & 4.5\% \\
\bottomrule
\end{tabular}%
}
\caption{\textbf{Performance and efficiency analysis on real-world physical data.} PFS improves mAP under both day-time and night-time conditions. While the additional 3.3~M parameters slightly increase latency, the module remains a lightweight add-on.}
\label{tab:performance_efficiency}
\end{table}

We evaluate the same frozen host detectors used in simulation on the collected 200 synchronized frames without additional fine-tuning. We compute class-wise AP for each object category and average across classes to obtain mAP. Table~\ref{tab:performance_efficiency} summarizes quantitative detection performance under daytime and nighttime conditions.

For BEVFusion, the baseline model achieves 32.87 mAP during daytime and 24.08 mAP under low-illumination nighttime conditions. Attaching PFS consistently improves performance in both regimes, reaching 35.33 mAP during the day and 29.20 mAP at night. This corresponds to a +2.46 mAP gain in daytime scenes and a larger +5.12 gain in nighttime scenes.

For UniBEV, the baseline detector achieves 34.17 mAP during daytime and 25.63 mAP at night. With PFS attached, performance improves to 36.85 mAP in daytime conditions and 30.41 mAP at night, corresponding to gains of +2.68 and +4.78 mAP, respectively.

Across both host detectors, PFS consistently improves detection accuracy without modifying the upstream image backbone, LiDAR encoder, or fusion architecture. Notably, performance gains are larger under low-illumination nighttime conditions, aligning with robustness trends observed under synthetic degradations in simulation.

It is important to note that the collected physical dataset utilizes a reduced sensor configuration consisting of a single forward-facing camera and a single 32-beam LiDAR, in contrast to the full surround multi-camera and high-resolution LiDAR setup available in the nuScenes benchmark. This limited field-of-view and sensing redundancy effectively introduces a partial observability condition analogous to structured modality dropout. As a result, absolute detection performance is lower than large-scale benchmark settings. Nevertheless, the consistent cross-detector improvements demonstrate that PFS generalizes beyond synthetic corruption suites and remains effective under real-world reduced-sensor deployments.

\subsubsection{Efficiency and Overhead}
We measure single-sample inference latency and throughput on an RTX 4080 Super GPU. The base detector remains frozen during evaluation. Table~\ref{tab:performance_efficiency} reports parameter count, latency, and FPS for BEVFusion and UniBEV with and without PFS. When attached to BEVFusion, PFS increases the parameter count from 40.8M to 44.1M and latency from 70.9 ms to 76.6 ms, corresponding to an 8.1\% runtime overhead while maintaining 13.1 FPS. For UniBEV, PFS increases latency from 417 ms to 436 ms with a 4.5\% overhead. These results confirm that PFS introduces limited computational overhead relative to the host detector while preserving real-time feasibility for deployment.


\section{CONCLUSIONS and FUTURE WORK}
\label{section:Conclusions}
\subsection{Discussion}
In this paper, we introduced PFS, a post-fusion BEV feature stabilization module that improves the robustness of BEV-based camera LiDAR fusion detectors under structured corruptions. PFS is lightweight for deployment, operates directly on the fused BEV features, and outputs a corrected representation to the original detection head without changing encoders, fusion design, or the head itself. It targets three common failure modes in fused BEV space global distribution shift, spatially localized corruption, and residual degradation by combining shift normalization, spatial reliability estimation, and gated residual correction, with identity initialization and frozen-backbone training to preserve clean performance. Experiments on nuScenes show that PFS maintains competitive clean accuracy while improving robustness under diverse camera and LiDAR corruptions, suggesting post-fusion stabilization as a practical, architecture-compatible route to more reliable multimodal 3D perception.

\subsection{Future Work}
While PFS provides a robust stabilization layer, future work will extend it along three directions. First, we will explore test-time adaptation so the shift normalization and reliability modules can update online under unseen domain shifts, using the Anchor Loss as a self-supervised signal to recalibrate reliability without labels~\cite{sun2020testtimetrainingselfsupervisiongeneralization}. Second, we will move from a single spatial reliability mask to channel-wise reliability prediction, enabling selective suppression of only the feature channels affected by specific sensor failures, inspired by channel-prediction and attention methods~\cite{mohsin2025channelpredictionnetworkdistribution}. Finally, we will incorporate temporal context with a recurrent or transformer-based aggregator so Expert Correction can leverage past frames to inpaint missing detections and improve robustness to transient dropouts and intermittent occlusions in dynamic scenes.


\bibliographystyle{IEEEtran}
\bibliography{IEEEabrv,ref}

\end{document}